\documentclass[runningheads]{llncs}
\usepackage{graphicx}

\usepackage{tikz}
\usepackage{comment}
\usepackage{amsmath,amssymb} 
\usepackage{color}

\usepackage[accsupp]{axessibility}  

\usepackage{booktabs} 

\usepackage{algorithm} 
\usepackage{listings}
\usepackage{epsfig}
\usepackage[title]{appendix}
\usepackage[caption=false]{subfig}
\usepackage{booktabs} 
\usepackage{tabulary}
\usepackage{tabulary,multirow,overpic,xcolor}
\newcolumntype{x}[1]{>{\centering\arraybackslash}p{#1pt}}

\newlength\savewidth
\newcommand{\tablestyle}[2]{\setlength{\tabcolsep}{#1}\renewcommand{\arraystretch}{#2}\centering\footnotesize}
\usepackage[colorlinks,linkcolor=black,bookmarks=false]{hyperref}
\definecolor{bananayellow}{rgb}{1.0, 0.88, 0.21}
\usepackage{wrapfig}

\newcommand{\netname}{EdgeViT}
\newcommand{\netsname}{EdgeViTs}
\newcommand{\modulename}{\textit{Local-Global-Local}}
\newcommand{\shortmodulename}{LGL}

\newcommand{\ie}{\textit{i.e.}}
\definecolor{amber}{rgb}{1.0, 0.49, 0.0}
\begin{document}
\pagestyle{headings}
\mainmatter
\def\ECCVSubNumber{3769}  

\title{\netsname: Competing Light-weight CNNs on Mobile Devices with Vision Transformers} 


\titlerunning{\netsname: Efficient Transformers}
%
\author{Junting Pan\inst{1}\thanks{Work done during an internship at Samsung AI Cambridge.}, Adrian Bulat \inst{2}, Fuwen Tan\inst{2}, Xiatian Zhu\inst{2}, Lukasz Dudziak\inst{2}, Hongsheng Li\inst{1}, Georgios Tzimiropoulos\inst{2,3} \and Brais Martinez \inst{2}
}
\authorrunning{Pan. et al.}
%
\institute{The Chinese University of Hong Kong \and Samsung AI Cambridge \and Queen Mary University of London}
\maketitle

\begin{abstract}
Self-attention based models such as vision transformers (ViTs) have emerged as a very competitive architecture alternative to convolutional neural networks (CNNs) in computer vision. Despite increasingly stronger variants with ever higher recognition accuracies, due to the quadratic complexity of self-attention, existing ViTs are typically demanding in computation and model size. Although several successful design choices (e.g., the convolutions and hierarchical multi-stage structure) of prior CNNs have been reintroduced into recent ViTs, they are still not sufficient to meet the limited resource requirements of mobile devices. This motivates a very recent attempt to develop light ViTs based on the state-of-the-art MobileNet-v2, but still leaves a performance gap behind. In this work, pushing further along this under-studied direction we introduce 
{\bf \netsname}, a new family of light-weight ViTs that, for the first time, enable attention based vision models to compete with the best light-weight CNNs in the tradeoff between accuracy and on-device efficiency. This is realized by introducing a highly cost-effective {\em local-global-local} (LGL) information exchange bottleneck based on optimal integration of self-attention and convolutions. For device-dedicated evaluation, rather than relying on inaccurate proxies like the number of FLOPs or parameters, we adopt a practical approach of focusing directly on on-device latency and, for the first time, energy efficiency. Extensive experiments on image classification, object detection and semantic segmentation validate high efficiency of our \netsname~when compared to the state-of-the-art efficient CNNs and ViTs in terms of accuracy-efficiency tradeoff on mobile hardware. Specifically, we show that our models are Pareto-optimal when both accuracy-latency and accuracy-energy tradeoffs are considered, achieving strict dominance over other ViTs in almost all cases and competing with the most efficient CNNs. Code is available at
\url{https://github.com/saic-fi/edgevit}.

\end{abstract}

\section{Introduction}
\label{sec:intro}

\begin{figure}
    \centering
    \includegraphics[width=0.9\linewidth]{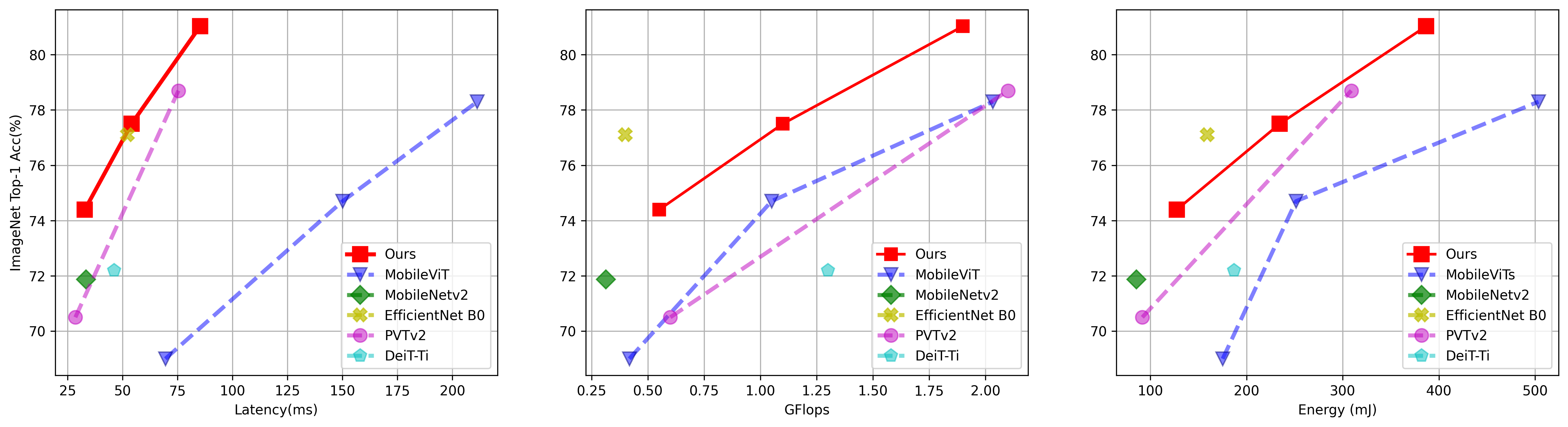}
    \caption{Our \netsname~yield comparable or superior tradeoff between accuracy and efficiency (e.g., run speed, latency) on ImageNet-1K, in comparison to state-of-the-art 
    efficient CNNs \cite{mobilenetv2_cvpr18,tan2019efficientnet},
    representative generic ViTs \cite{wang2021pvtv2,deit2021} and latest MobileViTs \cite{mehta2021mobilevit}. 
    Note that, \netsname{} outperform MobileViTs, the only specially designed model for mobile device, across all the metrics by a large margin.
    Whilst our lightest \netname~consumes more GFLOPs than EfficientNet B0 \cite{tan2019efficientnet}, it runs faster (33.3 vs. 52.1 ms per image), with a reduced gap in {\em on-device energy consumption}.
    {\em Testing device}: Samsung Galaxy S21 (latency), Snapdragon 888 Hardware Development Kit (energy).
    }
    \label{fig:fig1}
\end{figure}

Vision transformers (ViTs) have rapidly superseded convolutional neural networks (CNNs)
on a variety of visual recognition tasks \cite{vit2021,deit2021}, particularly when the priors and successful designs of previous CNNs are reintroduced for leveraging the induction bias of visual data such as local grid structures \cite{pvt,liu2021swin,chu2021twins,han2021transformer}.
Due to the quadratic complexity of ViTs and the high-dimension property of visual data,
it is indispensable that the computational cost needs to be taken into account in design \cite{pvt}.
Three representative designs to make computationally viable ViTs
are (1) the use of a hierarchical architecture with the spatial resolution (i.e., the token sequence length) progressively down-sampled across the stages \cite{mvit2021,pvt,chu2021twins},
(2) the locally-grouped self-attention mechanisms for controlling the length of input token sequences and parameter sharing \cite{liu2021swin,han2021transformer},
(3) the 
pooling attention schemes to subsample the \texttt{key} and \texttt{value} by a factor \cite{wang2020linformer,lu2021soft,xiong2021nystr,pvt}.
The general trend has been on designing more complicated and stronger ViTs to challenge
the dominance of top-performing CNNs \cite{resnet2016,regnet,tan2019efficientnet} in computer vision by achieving ever higher accuracies \cite{touvron2021going,zhou2021deepvit}.
These advances however are still insufficient to satisfy the design requirements and constraints for mobile and edge platforms
(e.g., smart phones, robotics, self-driving cars, AR/VR devices),
where the vision tasks need to carry out in a timely manner under certain computational budgets.
Prior efficient CNNs (e.g., MobileNets \cite{mobilenetv1,mobilenetv2_cvpr18,mobilenetv3_iccv19}, ShuffleNets \cite{ma2018shufflenet,zhang2018shufflenet}, EfficientNets \cite{tan2019efficientnet,tan2020efficientdet,tan2021efficientnetv2}, and etc.)
remain the state-of-the-art network architectures for such platforms in the tradeoff between running latency and recognition accuracy (Fig. \ref{fig:fig1}).

In this work, we focus on the development of largely under-studied {\em efficient ViTs} 
with the aim to surpass the CNN counterparts on mobile devices.
We consider a collection of very practical design requirements for running a ViT model on a target real-world platform as follows:
{\bf (1)} {\em Inference efficiency} needs to be high (e.g., {\em low latency and energy consumption}) so that
the running cost becomes generically affordable and more on-device applications 
can be supported.
This is a direct metric that we really care about in practice.
In contrast, the often-used efficiency metric, FLOPs (i.e., the number of multiply-adds),
cannot directly translate into the latency and energy consumption on a specific device, with several conditional factors including memory access cost, degree of parallelism, and the platform's characteristics \cite{ma2018shufflenet}.
This is, {\em not} all operations of a model can be carried out 
at the same speed and energy cost on a device.
Hence, FLOPs is merely an approximate and indirect metric of efficiency.
{\bf (2)} {\em Model size} (i.e., parameter number) is affordable for modern average devices.
Given the availability of ever cheaper and larger storage spaces,
this constraint has been relaxed significantly.
For example, an average smart phone often comes with 32GB or more storage.
As a consequence, using it as a threshold metric is no longer valid 
in most cases.
{\bf (3)}
{\em Implementational friendliness} is also critical in real-world applications.
For a wider range of deployment, it is necessary that a model can be implemented efficiently using the standard computing operations supported and optimized in the generic deep learning frameworks (e.g., ONNX, TensorRT, and TorchScript), without costly per-framework specialization. Otherwise, the on-device speed of a model might be unsatisfactory even with low FLOPs.
For instance, the cyclic shift and its reverse operations introduced in Swin Transformers \cite{liu2021swin} are rarely supported by the mainstream frameworks, i.e., deployment unfriendly. 
In the literature, very recent MobileViTs \cite{mehta2021mobilevit} are the only series of ViTs designed for mobile devices. In architecture design, they are a straightforward combination of MobileNetv2 \cite{mobilenetv2_cvpr18} and ViTs \cite{vit2021}.
As a very initial attempt in this direction, MobileViTs still lag behind CNN counterparts. 
Further, its evaluation protocol takes the {\em model size} (i.e. the parameter number) as the competitor selection criteria (i.e., comparing the accuracy of models only with similar parameter numbers), which however is no longer a hard constraint with modern hardware as discussed above and is hence out of date.

We present a family of light-weight attention based vision models, dubbed as {\bf \netsname},
for the first time, enabling ViTs to
compete with the best light-weight CNNs (e.g., MobileNetv2 \cite{mobilenetv2_cvpr18} and EfficientNets \cite{tan2019efficientnet})
in terms of accuracy-efficiency tradeoff on mobile devices.
This sets a milestone in the landscape of light-weight ViTs {\em vs.} CNNs in the low resource regime.
Our \netsname~are based on a novel factorization of
the standard self-attention for more cost-effective information exchange within every individual layer.
This is made possible by introducing a highly light-weight and easy-to-implement {\em local-global-local} (LGL) information exchange bottleneck
characterized with three operations:
{\bf(i)} Local information aggregation from neighbor {tokens} (each corresponding to a specific patch) using efficient depth-wise convolutions;
{\bf(ii)} Forming a sparse set of evenly distributed {\em delegate tokens} for long-range information exchange by self-attention;
{\bf(iii)} Diffusing updated information from {\em delegate tokens}
to the {\em non-delegate tokens} in local neighborhoods via transposed convolutions.
As we show in experiments, this design presents a favorable hybrid of self-attention,
convolutions, and transposed convolutions,
achieving the best accuracy-efficiency tradeoff.
It is efficient in that the self-attention is applied to a sparse set of delegate tokens.
To support a variety of computational budgets,
with our primitive module we establish a family of \netname~variants with three computational complexities: small (\texttt{S}), extra-small (\texttt{XS}), extra-extra-small (\texttt{XXS}).

We make the following \noindent{\bf contributions}: 
{\bf (1)} We investigate the design of light-weight ViTs from the practical on-device deployment and execution perspective.
{\bf (2)} For best scalability and deployment, we present a novel family of efficient ViTs, termed as \netsname, designed based 
on an optimal decomposition of self-attention using
standard primitive operations.
{\bf (3)} 
Regarding on-device performance, towards relevance for real-world deployment, we directly consider latency and energy consumption of different models rather than relying on high-level proxies like number of FLOPs or parameters.
Our results experimentally verify efficiency of our models in a practical setting and refute some of the claims made in the existing literature.
More specifically, extensive experiments on three visual tasks show that our \netsname~can match or surpass state-of-the-art light-weight CNNs, whilst consistently outperform the recent MobileViTs in accuracy-efficiency tradeoff,
including largely ignored on-device energy evaluation. 
Importantly, \netsname~are consistently Pareto-optimal in terms of both latency and energy efficiency, achieving strict dominance over other ViTs in almost all cases and competing with the most efficient CNNs.
On ImageNet classification our \netname-XXS outperforms MobileNetv2 by 2.2\% subject to the similar energy-aware efficiency.

\section{Related Work}
\label{sec:relatework}

{\flushleft \bf Efficient CNNs.} Since the advent of modern CNN architectures \cite{resnet2016,inceptionv3}, there has been a steady stream of works focusing on efficient architecture design for on-device deployment.
The first widely adopted families bring depthwise separable convolutions in a ResNet-like structure, e.g., MobileNets \cite{mobilenetv1,mobilenetv2_cvpr18}, ShuffleNets \cite{zhang2018shufflenet,ma2018shufflenet}. 
These works define a space of well-performing efficient architectures, resulting in widespread usage. 
Successive works further exploit this design space by automating the architectural design choices \cite{tan2019efficientnet,mobilenetv3_iccv19,tan2019mnasnet,tan2019mixconv}.
As a parallel line of research, net pruning creates efficient architectures by removing spurious parts of a larger network with close-to-zero weights \cite{deepcomp_iclr16,strucsparse_neurips16}, or via first training a super-network that is further slimmed to meet a pre-specified computational budget \cite{slimming_iccv17,Berman2020AOWS}.
Dynamic computing has also been explored, consisting of the mechanisms that condition the network parameters on the input data \cite{condconv_neurips19,dynamicconv_cvpr20}. 
Finally, using low bit-width is a very critical technique that can offer different tradeoffs between the accuracy and efficiency \cite{deepcomp_iclr16,Jacob_2018_CVPR,bitmixer_iccv21}.

{\flushleft \bf Vision transformers.}
ViTs \cite{vit2021} quickly popularize transformer-based architectures for computer vision. 
A series of works followed instantly, offering large improvements to the original ViTs in terms of data efficiency~\cite{deit2021,efficient_liu21} and architecture design \cite{liu2021swin,t2tvit,mvit2021,xvit_neurips21}. 
Among these works, one of the main modifications is to introduce hierarchical designs in multiple stages from convolutional architectures \cite{liu2021swin,chu2021twins,pvt,wang2021pvtv2}. 
Several works also focus on improving the positional encoding by using a relative positional embedding ~\cite{botnet2021,relposition_acl18}, making it learnable \cite{learnablepos_icml17}, or even replacing it by a attention bias element \cite{levit}.
All these approaches mostly aim to improve the model performance.

Recently, more efforts have been made towards finding efficient alternatives to the multi-head self-attention (MHSA) module, which is typically the computational bottleneck in the ViT architectures. 
A particularly effective solution is to reduce the internal spatial dimensions within the MHSA. 
The MHSA involves projecting the input tensor into key, query and value tensors. 
Several recent works, e.g. \cite{chu2021twins,pvt,wang2021pvtv2}, find that the key and value tensors could be downsampled with a limited loss in accuracy, leading to a better efficiency-accuracy tradeoff. 
Our work extends this idea by also downsampling the query tensors, which further improves the efficiency, as shown in Fig.~\ref{fig:flow}.

There are also alternative approaches reducing the number of tokens dynamically~\cite{rao2021dynamicvit,evovit2022,iared_neurips21,ats_arxiv21}.
That is, in the forward pass, tokens deemed to not contain the important information for the target task are pruned or pooled together, reducing the overall complexity thereafter. 
Finally, encouraged by their potential complementarity, many works have attempted to combine convolutional designs with self-attentions. This ranges from using convolutions at the stem \cite{xiao2021early}, integrating convolutional operations into the MHSA block \cite{uniformer_iclr22,cvt_iccv21}, or incorporating the MHSA block into ResNet-like architectures \cite{botnet2021}. It is interesting to note that even the original ViTs explored similar tradeoffs. \cite{vit2021}.

{\flushleft \bf Vision transformers for mobile devices.}
Whilst the efficiency issue has been taken into account in designing the ViT variants discussed above, they are still not dedicated and satisfactory architectures for on-device applications. 
There is only one exception, MobileViTs \cite{mehta2021mobilevit}, which are introduced very recently.
However, compared to the current best light-weight CNNs such as MobileNets \cite{mobilenetv2_cvpr18,mobilenetv3_iccv19} and EfficicentNets \cite{tan2019efficientnet}, these ViTs are still clearly inferior in terms of the on-device accuracy-efficiency tradeoff. 
In this work,
we present the first family of efficient ViTs that can deliver comparable or even superior tradeoffs in comparison to the best CNNs and ViTs.
We also extensively carry out the critical yet largely lacking on-device evaluations with energy consumption analysis.

\section{\netsname}
\label{sec:method}

\subsection{Overview}

\begin{figure}[t]
    \centering
    \includegraphics[width=0.9\linewidth]{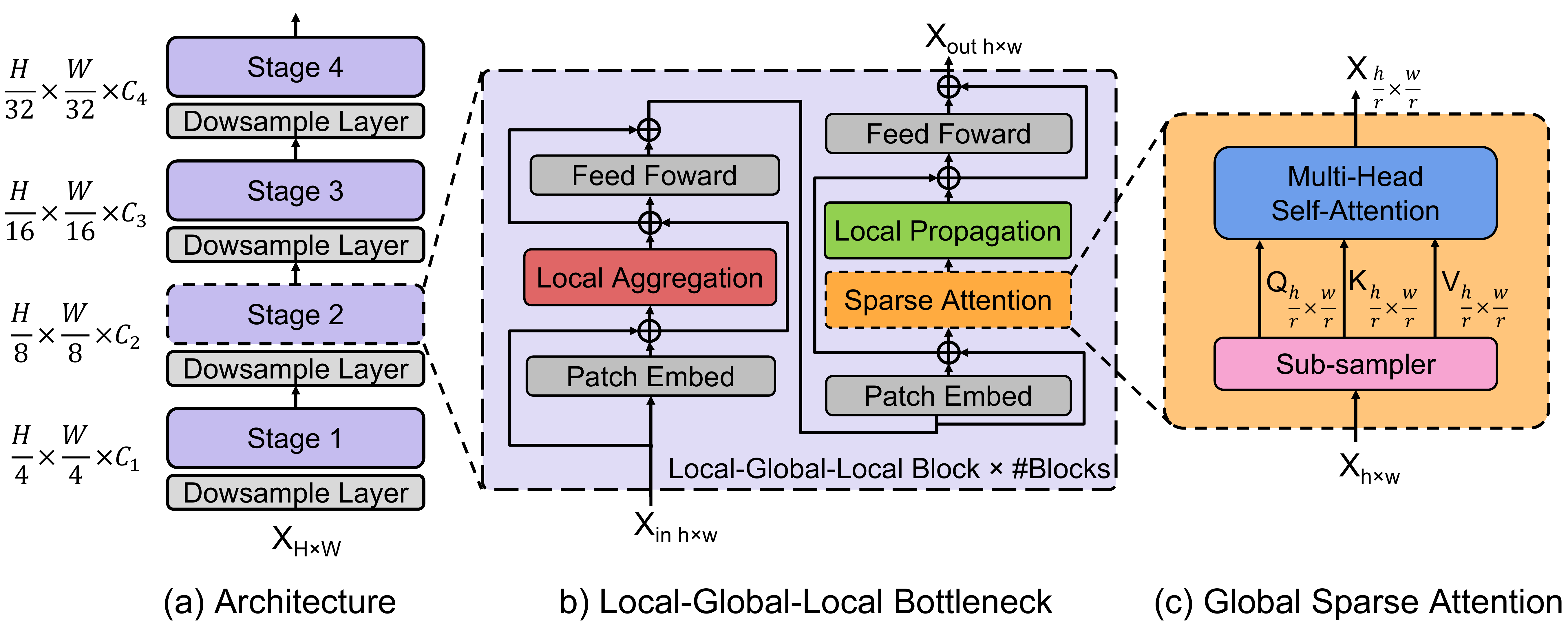}
    \caption{ (a) Schematic overview of our four stages \netname~architecture, with each stage consisting of a stack of 
    (b) 
    \modulename~(\shortmodulename) blocks constructed with local aggregation module,  sparse-self-attention and  local propagation module, patch embedding (PE) and Feed Forward Network (FFN). 
    In this example, ${h}$ and ${w}$ refer to input height and width of stage-2: $h=\frac{H}{8}$ and $w=\frac{W}{8}$. 
    $C_i$ refers to the number of channels for stage-$i$ and \textit{r} denotes the sub-sampling rate. 
    }
    \label{fig:flow}
\end{figure}

For designing light-weight ViTs suitable for mobile/edge devices,
we adopt a hierarchical pyramid network structure (Fig.~\ref{fig:flow}(a)) used in recent ViT variants \cite{pvt,cpvt,chu2021twins,wang2021pvtv2,mvit2021}.
A pyramid transformer model typically reduces the spatial resolution but expands the channel dimension across different stages.
Each stage consists of multiple transformer-based blocks processing tensors of the same shape, mimicking the ResNet-like networks.
The transformer-based blocks heavily rely on the self-attention operations at a quadratic complexity w.r.t the spatial resolution of the visual features.
By progressively aggregating the spatial tokens, pyramid vision transformers are potentially more efficient than isotropic models~\cite{vit2021}.
In this work, we dive deeper into the transformer-based block and introduce a cost-effective bottleneck, \modulename~(\shortmodulename) (Fig.~\ref{fig:flow}(b)).
\shortmodulename~further reduces the overhead of self-attention with a sparse attention module (Fig.~\ref{fig:flow}(c)), achieving better accuracy-latency balancing.

\begin{figure}[t]
    \centering
    \includegraphics[width=0.9\linewidth]{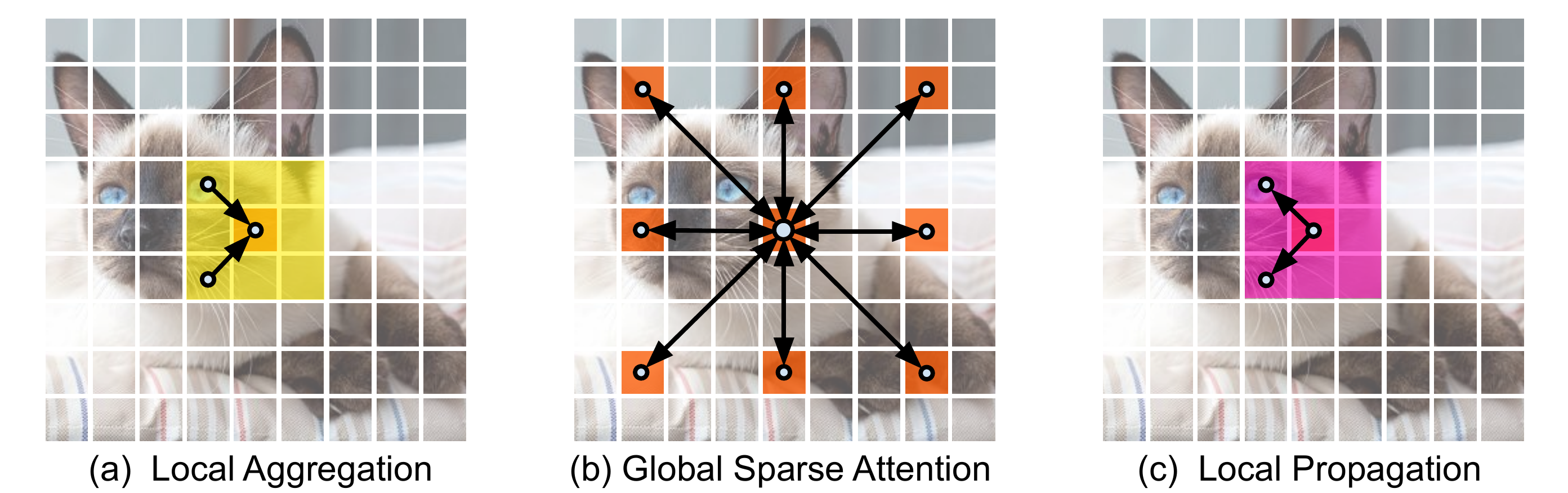}
    \caption{Illustration of three key operations involved in
    the proposed \modulename~(\shortmodulename) transformer block. In this example, we showcase how {\em the target token} (\textcolor{orange}{the orange square}) at the center conducts information exchange with all the others in three sequential steps:
    (a) Local information from neighbor tokens within the \textcolor{bananayellow}{yellow area} is first aggregated to the target token. (b) Global sparse attention is then computed among the target token and other selected delegates in orange color. (c) Global context information encoded in the target token is last propagated to its neighbor {\em non-delegate} tokens within the \textcolor{magenta}{pink area}.
    }
    \label{fig:self_att}
\end{figure}

\subsection{\modulename~bottleneck}
Self-attention has been shown to be very effective for learning the global context or long-range spatial dependency of an image, which is critical for visual recognition.
On the other hand, as images have high spatial redundancy (e.g., nearby patches are semantically similar)~\cite{mae}, 
applying attention to all the tokens, even in a down-sampled feature map, is inefficient.
There is hence an opportunity to reduce the scope of tokens whilst still preserving the underlying information flows that model the global and local contexts.
In contrast to previous transformer blocks that perform self-attention at each spatial location, 
our \shortmodulename~bottleneck only computes self-attention for a subset of the tokens but enables full spatial interactions, as in the standard multi-head self-attention (MHSA) \cite{vit2021}.

To achieve this, we decompose the self-attention into consecutive modules that process the spatial tokens within different ranges (Fig.~\ref{fig:flow}(b)). 
We introduce three efficient operations: i) \textit{\textbf{L}ocal aggregation} that integrates signals only from locally proximate tokens; ii) \textit{\textbf{G}lobal sparse attention} that model long-range relations among a set of delegate tokens where each of them is treated as a representative for a local window; iii) \textit{\textbf{L}ocal propagation} that diffuses the global contextual information learned by the delegates to the non-delegate tokens with the same window. 
Combining these, our \shortmodulename~bottleneck enables information exchanges between any pair of tokens in the same feature map at a low-compute cost. 
Each of these components is described in detail below:

\begin{itemize}
    \item \textit{\textbf{L}ocal aggregation}: 
    for each token, we leverage depth-wise and point-wise convolutions to aggregate information in local windows with a size of $k \times k$ (Fig. \ref{fig:self_att}(a)). 
    \item \textit{\textbf{G}lobal sparse attention}:
    we sample a sparse set of {\em delegate tokens} distributed evenly across the space, one {\em token} for each $r \times r$ window. Here, $r$ denotes the sub-sample rate. We then apply self-attention on these selected tokens only (Fig. \ref{fig:self_att}(b)).
    This is distinct from all the existing ViTs \cite{chu2021twins,pvt,wang2021pvtv2} where all the spatial tokens are involved as queries in the self-attention computation.
    \item \textit{\textbf{L}ocal propagation}:
    We propagate the global contextual information encoded in the delegate tokens to their neighbor tokens by transposed convolutions (Fig. \ref{fig:self_att}(c)).
\end{itemize}

\noindent Formally, our \shortmodulename~bottleneck can be formulated as:
\begin{equation}
    \begin{aligned}
    X &=\texttt{LocalAgg}(\texttt{Norm}(X_{in})) + X_{in},\\
    Y &= \texttt{FFN}( \texttt{Norm}(X) ) + X, \\
    Z &= \texttt{LocalProp}(\texttt{GlobalSparseAttn}(\texttt{Norm}(Y))) + Y, \\
    X_{out} &= \texttt{FFN}(\texttt{Norm}(Z)) + Z.\\
    \end{aligned}
\end{equation}

\noindent Here $X_{in} \in \mathcal{R}^{H\times W \times C}$ indicates the input tensors. 
$\texttt{Norm}$ is the layer normalization operation\cite{layernorm2016}.
$\texttt{LocalAgg}$ represents the local aggregation operator,
$\texttt{FFN}$ is a two-layer perceptron, similar to the position-wise feed-forward network introduced in~\cite{vit2021}.
$\texttt{GlobalSparseAttn}$ is the global sparse self-attention.
$\texttt{LocalProp}$ is the local propagation operator.
For simplicity, positional encoding is omitted. 
Note that, all these operators can be implemented by commonly used and highly optimized operations in the standard deep learning platforms.
Hence, our LGL bottleneck is implementation friendly.

{\flushleft \bf Comparisons to existing designs.} 
Our~\shortmodulename~bottleneck shares a similar goal with the recent PVTs \cite{pvt,wang2021pvtv2} and Twins-SVTs \cite{chu2021twins} models that attempt to reduce the self-attention overhead. However, they differ in the core design.
PVTs \cite{pvt,wang2021pvtv2} perform self-attention where the number of \texttt{keys} and \texttt{values} are reduced by strided-convolutions, whilst the number of \texttt{queries} remains the same. 
In other words, PVTs still perform self-attention at each grid location.
In this work, we question the necessity of positional-wise self-attention and explore to what extent the information exchange enabled by our LGL bottleneck could approximate the standard MHSA (see Section \ref{sec:experiments} for more details).
Twins-SVTs \cite{chu2021twins} combine local-window self-attention~\cite{liu2021swin} 
with global pooled attention from PvTs~\cite{pvt}.
This is different from our hybrid design using both self-attention
and convolution operations distributed in a series of {\em local-global-local} operations.
As demonstrated in the experiments (Table \ref{tab:imagenet_main} and \ref{tab:imagenet_energy}),
our design achieve a better tradeoff between the model performance and the computation overhead (e.g. latency, energy consumption, etc).

\begin{table}[!t]
\centering
\scalebox{1}{
\tablestyle{4.8pt}{1.1}
\begin{tabular}{@{}l|x{90}|x{35}|x{35}|x{35}|x{35}@{}}
 Model & \#Channels & \#Blocks & \#Heads &FLOPs & \#Param\\
\toprule
 {\netname-XXS} & [36, 72, 144, 288] & [1,1,3,2] & [1,2,4,8] & 0.56G & 4.1M \\
 {\netname-XS }& [48, 96, 240, 384] & [1,1,2,2] & [1,2,4,8] & 1.1G & 6.7M \\
 {\netname-S} & [48, 96, 240, 384] & [1,2,3,2] & [1,2,4,8] & 1.9G & 11.1M\\
 \bottomrule
\end{tabular}
}
\caption{\textbf{Configuration of three \netname~variants.}
``\#Channels'': number of channels per stage.
``\#Blocks'': number of \shortmodulename~blocks per stage.
``\#Heads'': number of attention heads in MHSA.
``\#Param'': number of parameters.
\label{tab:variants}}
\end{table}

\subsection{Architectures}

We build a family of \netsname~with the proposed \shortmodulename~bottleneck at different computational complexities (i.e. 0.5G, 1G, and 2G).
The configurations are summarized in Table \ref{tab:variants}.
Following the hierarchical ViTs \cite{chu2021twins,pvt,liu2021swin,uniformer_iclr22}, 
\netsname~consist of four stages with the spatial resolution (\ie, the token sequence length) gradually reduced throughout, and their self-attention module replaced with our \shortmodulename~bottleneck. 
For the stage-wise down-sampling, we use a conv-layer with a kernel size of $2\times2$ and stride 2, except for the first stage where we down-sample the input feature by $\times4$, and use a $4\times4$ kernel and a stride of 4.
We adopt the conditional positional encoding \cite{cpvt}
that has been shown to be superior to the absolute positional encoding.
This can be implemented using 2D depth-wise convolutions with a residual connection.
In our model, we use $3\times3$ depth-wise convolutions with zero paddings.
It is placed before the local aggregation and global sparse self-attention.
The \texttt{FFN} consists of two linear layers with GeLU non-linearity \cite{gelu} placed in-between.
Our local aggregation operator is implemented as a stack of pointwise and depthwise convolutions.
The global sparse attention is composed of a spatial uniform sampler with sample rates of $(4, 2, 2, 1)$ for the four stages, and a standard MHSA. The local propagation is implemented with a depthwise separable transposed convolution with the kernel size and stride equal to the sample rate used in the global sparse attention. 
The exact architecture for the LGL bottleneck is described in the \texttt{supplementary material}.

\section{Experiments}
\label{sec:experiments}

We benchmark \netsname~on visual recognition tasks. 
We pre-train \netsname~on the Imagenet1K recognition task~\cite{imagenet2015}, comparing the performances and computation overheads against alternative approaches.
We also evaluate the generalization capacity of \netsname~on downstream dense prediction tasks: object detection and instance segmentation on the COCO benchmark~\cite{coco2014}, and semantic segmentation on the ADE20K Scene Parsing benchmark~\cite{ADE20K}.
For on-device execution, we report exucution time (latency) and energy consumption of all relevant models on ImageNet.
We do not report on-device measurements on downstream tasks as they reuse ImageNet models.

\subsection{Image Classification on ImageNet-1K}

{\flushleft \bf Training Settings. }
ImageNet-1K \cite{imagenet2015} provides 1.28 million training images and 50,000 validation images from 1000 categories.
We follow the training recipe introduced in DeiT~\cite{deit2021}.
We optimize the models using AdamW \cite{adamw} with a batch size of 1024, weight decay of $5\times10^{-2}$, and momentum of 0.9.
The models are trained from {\em scratch} for 300 epochs with a linear warm-up during the first 5 epochs.
Our base learning rate is set as $1\times10^{-3}$, and decay after the warm-up using a cosine schedule \cite{loshchilov2016sgdr}.
We apply the same data augmentations as in \cite{deit2021,chu2021twins,wang2021pvtv2,uniformer_iclr22} which include random cropping, random horizontal flipping, mixup, random erasing and label-smoothing. 
During training, the images are randomly cropped to $224\times224$.
During testing, we use a single center crop of $224\times224$.
We report the top-1 accuracy on the validation set.

{\flushleft \bf Benchmarking Settings. }
For latency measurements, we use a Samsung Galaxy S21 mobile phone equipped with a Snapdragon 888 chipset.
All relevant models are benchmarked by running a forward pass 50 times using TorchScript lite interpreter via the Android benchmarking app provided by PyTorch \cite{pytorch}.
We use CPU implementation, full precision and batch 1 to execute all operations.
This choice comes from the fact that this is the only combination that was able to robustly execute all of the models from our paper.
In general, more efficient implementations exist, such as those utilizing specialized hardware like Neural Processing Units (NPUs).
However, these put more restrictions on what can be executed and many models failed to run in our experiments when trying to use different hardware targets.

For energy measurements, we use a Monsoon High Voltage Power Monitor connected to a Snapdragon 888 Hardware Development Kit (HDK8350) to obtain accurate power readings over the course of running a forward pass of each test model 50 times.
The same TorchScript runtime is used as in latency measurements.
From the power signal reported by the monitor, we derive the average per-inference power and energy consumption by first subtracting background power consumption (i.e., power readings when not running any model) and then identifying 50 continuous regions of significantly higher power draw.
Each region like that is considered a single inference and we calculate its total energy as the integral over the individual power samples. Analogously we also calculate average power consumption by averaging over the same set of samples.
After energy and power are calculated for each inference, the final statistics of a model are obtained by again averaging over the 50 identified runs.
Our methodology follows what can be found in the literature~\cite{smartatwhatcost}.

\begin{table}[!t]
\centering
\scalebox{0.9}{
\tablestyle{4.8pt}{1.1}
\tablestyle{10pt}
{1.1}\begin{tabular}{@{}l|x{37}|x{45}|x{45}|x{60}@{}}
\toprule
 Model & \#Params & FLOPs & CPU(ms) & Acc Top-1 (\%) \\
\toprule
 {MobileNet-v2}\cite{mobilenetv2_cvpr18} & 3.4M & 0.3G & 33.3{\tiny$\pm$5.3}  & 72.0 \\
 {\bf MobileNet-v3 0.75}\cite{mobilenetv3_iccv19} & \bf 4.0M & \bf 0.16G & \bf{23.0{\tiny$\pm$3.7}} & \bf 73.3 \\
 \bf {EfficientNet-B0}\cite{tan2019efficientnet} & \bf 5.3M & \bf 0.4G & \bf 52.1{\tiny$\pm$7.4} & \bf 77.1 \\
  \midrule
 {MobileViT-XXS} \cite{mehta2021mobilevit}& 1.3M &  0.4G & 69.5{\tiny$\pm$5.1} & 69.0\\
 {PVT-v2-B0}\cite{wang2021pvtv2} & 3.4M & 0.6G & 26.0{\tiny$\pm$6.9} & 70.5\\
 {Uniformer-Tiny*}\cite{uniformer_iclr22} & 3.9M & 0.6G & 40.5{\tiny$\pm$3.1} & 74.1\\
 {Twins-SVT-Tiny*}\cite{chu2021twins} & 4.1M & 0.6G & 36.9{\tiny$\pm$2.3} & 71.2\\
 {\bf \netname-XXS} & \bf 4.1M & \bf 0.6G & \bf 32.8{\tiny$\pm$2.7} & \bf 74.4 \\
 \midrule
 {T2T-ViT-7} \cite{t2tvit}& 4.3M & 1.1G & 48.8{\tiny$\pm$6.5} &71.7 \\
 {MobileViT-XS} \cite{mehta2021mobilevit}& 2.4M & 1.1G & 150.1{\tiny$\pm$6.1} & 74.7\\
 {DeiT-Tiny} \cite{deit2021}& 5.7M & 1.3G &  46.2{\tiny$\pm$13.6} & 72.2\\
 {TNT-Tiny} \cite{han2021transformer}& 6.1M & 1.4G & 86.4{\tiny$\pm$6.0} &73.9 \\
 {\bf \netname-XS} & \bf 6.7M & \bf 1.1G & \bf 54.1{\tiny$\pm$2.2} & \bf 77.5 \\
 \midrule
 {T2T-ViT-12} \cite{t2tvit}& 6.9M & 1.9G & 69.9{\tiny$\pm$5.6} &76.5 \\
 {PVT-v2-B1}\cite{wang2021pvtv2} & 14M & 2.1G & 75.4{\tiny$\pm$2.3} & 78.7\\
 {MobileViT-S} \cite{mehta2021mobilevit}& 5.6M & 2.0G & 221.3{\tiny$\pm$9.3} & 78.3\\
  \bf {LeViT-384}$\dagger$ \cite{levit} &\bf 39.1M & \bf 2.4G & \bf 71.3{\tiny$\pm$2.3} & \bf 79.5\\
 {\bf \netname-S} & \bf 11.1M & \bf 1.9G & \bf 85.3{\tiny$\pm$3.9} & \bf{81.0} \\
 \hline
 
\end{tabular}

}
\caption{\textbf{Results on ImageNet-1K validation set.} All models are tested on input scale of $224\times224$, except for MobileViTs \cite{mehta2021mobilevit} that are tested with $256\times256$ according to their original implementation. * indicates down-scaled architectures beyond original definitions by authors to fit the mobile compute budget. {LeViT-384}$\dagger$ denotes the LeViT model retrained under the same setting as our EdgeViT. \label{tab:imagenet_main}}
\end{table}

\begin{table}[!t]
\centering
\scalebox{0.9}{
\tablestyle{4.8pt}{1.1}
\begin{tabular}{@{}l|x{45}|x{45}|x{50}|x{50}|x{45}@{}}
\toprule
Model & Top-1 (\%) & CPU (ms) & Energy (mJ) & Power\newline (W) & Efficiency (\%/msW) \\
\toprule
{MobileNet-v2}\cite{mobilenetv2_cvpr18}       & 72.0       & 33.3       & 85.7{\tiny$\pm$7.4}         & 3.31{\tiny$\pm$0.26}       & 0.841       \\
{MobileNet-v3 0.75}\cite{mobilenetv3_iccv19}  & {\bf 73.3} & {\bf 23.0} & {\bf 63.0{\tiny$\pm$9.6}}   & {\bf 3.46{\tiny$\pm$0.4}}  & {\bf 1.164} \\
{EfficientNet-B0}\cite{tan2019efficientnet}   & {\bf 77.1} & {\bf 52.1} & {\bf 159.0{\tiny$\pm$26.2}} & {\bf 3.62{\tiny$\pm$0.45}} & {\bf 0.485} \\
\midrule
{PVT-v2-B0}\cite{wang2021pvtv2}              & 70.5       & 26.0       & 91.7{\tiny$\pm$19.7}        & 3.94{\tiny$\pm$0.68}       & 0.769       \\
{PVT-v2-B1}\cite{wang2021pvtv2}              & {\bf 78.7} &  75.4 & 309.0{\tiny$\pm$65.8} & 4.63{\tiny$\pm$0.71} & 0.255 \\
\midrule
{Twins-SVT-Tiny*}\cite{chu2021twins}             & 71.2       & 36.9       & 114.5{\tiny$\pm$17.3}       & 3.71{\tiny$\pm$0.24}       & 0.622       \\
{DeiT-Tiny} \cite{deit2021}       & 72.2       & 46.2       & 187.2{\tiny$\pm$7.6}        & 4.77{\tiny$\pm$0.21}       & 0.386       \\
{Uniformer-Tiny*}\cite{uniformer_iclr22}     & 74.1       & 40.5       & 134.7{\tiny$\pm$27.3}       & 4.1{\tiny$\pm$0.71}        & 0.55        \\
{T2T-ViT-12} \cite{t2tvit}           & 76.5       & 69.9       & 266.2{\tiny$\pm$42.6}       & 4.37{\tiny$\pm$0.36}       & 0.287       \\
{TNT-Tiny} \cite{han2021transformer}         & 73.9       & 86.4       & 308.7{\tiny$\pm$70.5}       & 3.94{\tiny$\pm$0.63}       & 0.239       \\
{LeViT-384}$\dagger$ \cite{levit} & \bf 79.5  &  \bf 71.3{\tiny$\pm$2.2}& \bf 455.2{\tiny$\pm$125.8} & \bf 6.18{\tiny$\pm$0.74} & \bf 0.173  \\
\midrule
{MobileViT-XXS} \cite{mehta2021mobilevit}    & 69.0       & 69.5       & 175.3{\tiny$\pm$28.7}       & 2.77{\tiny$\pm$0.24}       & 0.394       \\
{MobileViT-XS} \cite{mehta2021mobilevit}     & 74.7       & 150.1      & 251.5{\tiny$\pm$81.1}       & 2.63{\tiny$\pm$0.61}       & 0.297       \\
{MobileViT-S} \cite{mehta2021mobilevit}      & 78.3       & 221.3      & 503.6{\tiny$\pm$117.0}      & 2.76{\tiny$\pm$0.21}       & 0.155       \\
\midrule
{\bf \netname-XXS}         & {\bf 74.4} & {\bf 32.8} & {\bf 127.4{\tiny$\pm$27.3}} & {\bf 4.27{\tiny$\pm$0.67}} & {\bf 0.584} \\
{\bf \netname-XS}          & {\bf 77.5} & {\bf 54.1} & {\bf 234.6{\tiny$\pm$44.0}} & {\bf 4.77{\tiny$\pm$0.84}} & {\bf 0.33}  \\
{\bf \netname-S}           & {\bf 81.0} & {\bf 85.3} & {\bf 386.7{\tiny$\pm$43.5}} & {\bf 4.8{\tiny$\pm$0.26}}  & {\bf 0.209} \\
\bottomrule
\end{tabular}

}
\caption{\textbf{On-device energy evaluation on ImageNet-1K.} All relevant metrics are reported as mean values per forward pass across 50 executions. 
For facilitating comparison, we define an energy-aware {\em efficiency} metric as the average gain in top-1 accuracy from each 1W run for 1ms (equivalent to consuming 1mJ of energy).
({\em Pareto-optimal models} are highlighted in bold in the last column).
}
\label{tab:imagenet_energy}
\end{table}

{\flushleft \bf Results.} 
We compare \netsname~to a variety of baseline models, including the classic efficient CNNs, e.g. MobileNetV2 \cite{mobilenetv2_cvpr18}, MobileNetV3 \cite{mobilenetv3_iccv19}, EfficientNet \cite{tan2019efficientnet}, and the state-of-the-art ViTs, e.g. MobileViT~\cite{mehta2021mobilevit}, PVT-v2~\cite{wang2021pvtv2}, DeiT~\cite{deit2021}, LeViT~\cite{levit}.
As the original LeViT \cite{levit} was optimized in a large-scale setting (i.e. 1000 epochs) with knowledge distillation, we perform a comparison by re-training LeViT under the same setting (300 epochs) as EdgeViT without knowledge distillation. We denote the retrained LeViTs as {LeViT-384}$\dagger$. 
We select the baselines with a complexity of less than 2 GFLOPS as i) in real-world applications, the computational cost remains the top concern; ii) whilst FLOPs is an indirect metric for the latency, it is the most used cost metric in prior works. This selection criterion is different from \cite{mehta2021mobilevit} that instead uses the model size (i.e. the parameter number) which however has become a less restricted facet in mobile devices.

From Table \ref{tab:imagenet_main}, we can learn:
i) \netsname~significantly outperform other light-weight {\bf \em ViTs} at a similar level of GFLOPs complexity. 
Compared to the PVT-v2 family \cite{wang2021pvtv2}, our \netname-XXS/\netname-S achieve $3.9\%$/$2.3\%$ improvements over PVT-v2-B0/PVT-v2-B1. Compared to MobileViTs,\netsname~achieve $5.4\%$, $2.8\%$ and $2.7\%$ gains in the three complexitiy settings.
ii) {\bf \em ViTs vs. CNNs:} 
Our \netsname~lift the performance of efficient ViTs to approach the level of well-established efficient CNNs.
For example, the \netname-XXS performs superior to MobileNet-v2 and MobileNet-v3-0.75 at a similar level of model size, but requires more GFLOPs.
However, we observe that the efficient CNNs still surpass efficient ViTs in the accuracy-FLOPs tradeoff by a small margin.

On the other hand, as discuss early, numbers of FLOPs or parameters are merely indicative but do not fully reflect the on-device efficiency \cite{ma2018shufflenet,brpnas}.
We further consider on-device latency and energy consumption directly.
Other than the representative ViTs and CNNs, we also compare two recent ViT variants \cite{uniformer_iclr22,chu2021twins} with the number of channels and layers re-scaled to fit the complexity need. 
%
As presented in Table \ref{tab:imagenet_main},
\netsname~demonstrate strong performance with latencies comparable to MobileNets: 
\netname-XXS achieves a gain of $2.4\%$ over MobileNet-V2 while running slightly faster. 
\netname-XXS also surpasses MobileNet-V3 by $1.1\%$ but at the cost of being 9.8ms slower. 
\netname-XS performs on par with the auto-searched EfficientNet-B0 model. 
We believe our models could also benefit from the automatic architecture search techniques as use in MobileNet-V3 and EfficientNets.
Our models yield clear advantages over alternative ViT models. 
Compared to MobileViTs in the three GFLOPs settings, \netsname~excel by 5.4\%, 2.8\%, and 2.7\% while being $\times2, \times2.7, \times2.6$ faster.

Energy results are presented in Table~\ref{tab:imagenet_energy}.
In addition to the raw energy and power numbers, 
for comparison simplicity, we define an energy-aware {\bf\em efficiency} metric as the average gain in top-1 accuracy (in percentages) from each consumed 1mJ of energy.
We observe that less accurate models tends to be more efficient. This is not a surprise in that improvements in accuracy scale sublinearly with model complexity.
However, this also means that comparing efficiency of models with very different top-1 scores might be severely biased by the sole difficulty of achieving certain accuracy levels, which is independent from a model.
Therefore, we limit our comparison to identifying {\bf\em pareto-optimal models}, those upon which no other models can improve in either accuracy or energy efficiency without degrading other metrics.
We can see that our \netname~family is able to dominate almost all other ViTs, with the only exception being LeViT-384$\dagger$ whose accuracy and efficiency fall between our EdgeViT-S and EdgeViT-XS.
When compared to CNNs, our \netsname~compete with MobileNet-v3 and EfficientNet-B0 that are more efficient but also less accurate.
MobileNet-v2 achieves decent results but is dominated by its newer version, MobileNet-v3.
PVT-v2-B0, although high on the efficiency side, is rather inaccurate and hence is favored by highly efficient CNNs.
Visibly at the end of the spectrum are the latest MobileViT models which turn out to be neither efficient nor accurate, when compared to the rest.
Unlike them, our \netname~models, although not as efficient as best CNNs in the absolute sense, exhibit favourable trade-off between efficiency and accuracy by being rather highly accurate while not sacrificing efficiency.

\begin{table*}[!t]
\centering
\scalebox{1}{
\subfloat[\textbf{Local Aggregation}\label{tab:ablation:local}]{%
\tablestyle{4.8pt}{1.1}
\tablestyle{4.8pt}{1.1}\begin{tabular}{@{}l|x{30}|x{18}@{}}
 & CPU & Top1\\
\toprule 
 {w/o LA} & 33.9ms & 72.7 \\ 
 {LA(LSA)} & 36.1ms & 74.0 \\
 {LA(Ours)} & \bf 32.8ms& \bf 74.4\\
\end{tabular}}\hfill
\subfloat[\textbf{Global attention}\label{tab:ablation:sampler}]{%
\tablestyle{4.8pt}{1.1}
\tablestyle{4.8pt}{1.1}\begin{tabular}{@{}l|x{30}|x{18}@{}}
 & CPU & Top1\\
\toprule 
 {max} & 34.8ms & 74.3 \\
 {avg} & 34.5ms & \bf 74.5 \\
 {center} & \bf 32.8ms & 74.4\\
\end{tabular}}\hfill
\subfloat[\textbf{Local Propagation}\label{tab:ablation:lp}]{%
\tablestyle{4.8pt}{1.1}
\tablestyle{4.8pt}{1.1}\begin{tabular}{@{}l|x{30}|x{18}@{}}
 & CPU & Top1\\
\toprule 
 {w/o LP} & \bf 32.4ms & 73.9 \\
 {LP(Bilinear)} & 34.1ms & 74.1 \\
 {LP(Ours)} & 32.8ms & \bf 74.4 \\
\end{tabular}}\hfill
}
\caption{\textbf{Ablation on ImageNet-1K.}
\texttt{LA}: the local aggregation operator. 
\texttt{LP}: the local propagation operator. 
\texttt{LSA}: the Locally-grouped Self-Attention used in~\cite{chu2021twins}. 
}
\end{table*}

\subsubsection{Ablation study.}
We conduct detailed ablations to validate our design choices in the \shortmodulename~bottleneck. 
We use \netname-XXS as the base model and re-scale the alternative designs to $\sim$0.5GFLOPs for fair comparison. \\

\textit{\textbf{L}ocal aggregation.} 
We compare our local aggregation (LA) operation to the Locally-grouped Self-Attention (\texttt{LSA}) used in~\cite{chu2021twins,liu2021swin}.
It is shown in Table \ref{tab:ablation:local} that applying LA consistently improve the performance.
Our convolutional LA module performs better than the self-attention based operator (\texttt{LSA}). 
This validates our choice of using \texttt{depth-wise convolutions} in LA for local context learning.

\textit{\textbf{G}lobal sparse attention.} 
We explore three options for delegate token sampling: \texttt{max}, \texttt{avg}, and \texttt{center}. 
All choices perform similarly in terms of accuracy, with our default design \texttt{center} being slightly faster. 

\textit{\textbf{L}ocal propagation.}
We investigate two alternatives to the local propagation operator: 
i) \texttt{w/o LP}: We simply remove the local propagation. Note that \netsname w/o LP has similar complexity to standard \netsname.
%
ii) \texttt{Bilinear}: we use the bilinear interpolation, instead of the transposed convolution, to up-sample the delegate tokens.
Table \ref{tab:ablation:lp} shows that adding LP improves the top-1 accuracy by $\mathtt{0.5\%}$, with only $\mathtt{0.4 ms}$ overhead.


\begin{table*}[t]
\centering
\scalebox{0.85}{
	\footnotesize

\begin{tabular}{l|c|ccc|ccc|c|ccc|ccc}
\toprule
\multirow{4}{*}{Backbone} &\multicolumn{7}{c|}{RetinaNet 1$\times$} &\multicolumn{7}{c}{Mask R-CNN 1$\times$}\\ 
\cmidrule{2-15}  
& \#Par.& AP & AP$_{50}$ & AP$_{75}$ & AP$_S$ & AP$_M$ & AP$_L$ &\#Par. &AP$^{\rm b}$ &AP$_{50}^{\rm b}$ &AP$_{75}^{\rm b}$  &AP$^{\rm m}$ &AP$_{50}^{\rm m}$ &AP$_{75}^{\rm m}$ \\
\toprule 
PVTv2-B0~\cite{wang2021pvtv2} & \textbf{13.0} & 37.2 & 57.2 & 39.5 & \textbf{23.1} & 40.4 & 49.7 &\textbf{23.5} & 38.2 & 60.5 & 40.7 & 36.2 & 57.8 & 38.6\\  
\bf \netname-XXS  & 13.1 & \textbf{38.7} & \textbf{59.0} & \textbf{41.0} & 22.4 & \textbf{42.0} & \textbf{51.6} & 23.8 & \textbf{39.9} & \textbf{62.0} & \textbf{43.1} & \textbf{36.9} & \textbf{59.0} & \textbf{39.4}  \\
\midrule 
\bf \netname-XS  & 16.3 & 40.6 & 61.3 & 43.3 & 25.2 & 43.9 & 54.6 & 26.5 & 41.4 & 63.7 & 45.0 & 38.3 & 60.9 & 41.3  \\
\midrule 
ResNet18~\cite{resnet2016} &\textbf{21.3} & 31.8 & 49.6 & 33.6 & 16.3 & 34.3 & 43.2 &\textbf{31.2} & 34.0 & 54.0 & 36.7 & 31.2 & 51.0 & 32.7\\ 
PVTv1-Tiny~\cite{pvt} &23.0& {36.7}& {56.9}& {38.9}& {22.6}& {38.8} &{50.0}  &32.9 & {36.7} & {59.2} & {39.3} & {35.1} & {56.7} & {37.3} \\
PVTv2-B1~\cite{wang2021pvtv2} &23.8 & 41.2 & 61.9 & 43.9 & 25.4 & 44.5 & 54.3 &33.7 & 41.8 & 64.3 & 45.9 & 38.8 & 61.2 & 41.6\\
\bf \netname-S  & 22.6 & \textbf{43.4} & \textbf{64.9} & \textbf{46.5} & \textbf{26.9} & \textbf{47.5} & \textbf{58.1}  & 32.8 & \textbf{44.8} & \textbf{67.4} & \textbf{48.9} & \textbf{41.0} & \textbf{64.2} & \textbf{43.8}\\
\bottomrule
\end{tabular}
}
	\caption{Comparison to other visual backbones using RetinaNet and Mask-RCNN on COCO \texttt{val2017} object detection and instance segmentation. ``\#Par.'' refers to number of parameters in million. AP$^{\rm b}$ and AP$^{\rm m}$ indicate bounding box AP and mask AP.
	}
	\label{tab:det_table} 
\end{table*}

\subsection{Dense Prediction}
Following~\cite{pvt,wang2021pvtv2}, we also evaluate the proposed~\netsname~on COCO Objection Detection/Instance Segmentation~\cite{coco2014} and ADE20K Scene Parsing~\cite{ADE20K}. 
Here, we use the ~\netsname~as the feature extractor for the main model and initialize it with the ImageNet1K-pretrained weights obtained in our previous experiments.

\subsubsection{COCO Object Detection/Instance Segmentation.}
We demonstrate the performance of our model in main-stream object detection and instance segmentation frameworks: RetinaNet~\cite{retinanet} for object detection, Mask R-CNN\cite{maskrcnn} with the FPN~\cite{fpn2017} for instance segmentation.
Following the training protocol in~\cite{wang2021pvtv2,chu2021twins,uniformer_iclr22}, 
we resize the training images to have a shorter side of 800 pixels while keeping the longer side to be smaller than 1333 pixels.
During testing, the images are re-scaled to have a shorter size of 800 pixels.
The models are finetuned with $1\times$ schedule (i.e. 12 epochs) by AdamW\cite{adamw} using an initial learning rate of $1\times10^{-4}$ and a batch size of 16.
We train the models on the COCO 2017 training set and report the mAP@100 score on the COCO 2017 validation set.

{\flushleft \bf Results.} 
In Table~\ref{tab:det_table}, our \netsname~perform consistently better than other visual backbones on RetinaNet~\cite{retinanet} and Mask R-CNN\cite{maskrcnn}. 
Our smallest variant \netname-XXS, when used on RetinaNet~\cite{retinanet}, achieves $1.5$ higher AP than PVTv2-B0.
When used on Mask R-CNN~\cite{maskrcnn}, \netname-XXS also surpasses PVTv2-B0 by $1.7$ on the bounding box detection task (AP$^{\rm b}$), and by $0.7$ on the mask segmentation task (AP$^{\rm m}$). 
For \netname-S, we observe even larger gains when comparing to PVTv2-B1: $+2.2$ on RetinaNet~\cite{retinanet}, $+3.0$ AP$^{\rm b}$ and $+1.2$ AP$^{\rm m}$ on Mask R-CNN\cite{maskrcnn}. 

\subsubsection{ADE20K Scene Parsing.}
We incorporate the pretrained~\netname~ in the Semantic FPN segmentation model~\cite{semantic_fpn}.
As in~\cite{pvt,wang2021pvtv2}, we create $512\times512$ random crops of the images during training and resize the images to have a shorter side of 512 pixels during inference.
The models are finetuned by AdamW~\cite{adamw} using an initial learning rate of $1\times10^{-4}$ and a batch size of 16.
We train the models for 80K iterations on the ADE20K training set, and report the mean Intersection over Union (mIoU) score on the validation set.

{\flushleft \bf Results} 
In Table~\ref{tab:seg}, we compare \netsname~to both CNN (ResNet-18~\cite{resnet2016}) and ViT backbones (PVTs\cite{pvt,wang2021pvtv2}) for FPN based Semantic Segmentation~\cite{semantic_fpn}. 
\netsname~achieves better performance than all counterparts at similar compute costs. 
Particularly, \netname-XXS outperforms PVTv2-B0 by 2.5$\%$ in mIoU, \netname-S~surpasses PVTv2-B1 by a margin of 3.4$\%$.


\begin{table*}[t]
\centering
\scalebox{0.85}{
	\footnotesize
	\setlength{\tabcolsep}{0.8cm}
\begin{tabular}{l|c|c|c}
\toprule
\multirow{2}{*}{Backbone} & \multicolumn{3}{c}{Semantic FPN}\\
\cmidrule{2-4}
& \#Param (M) & GFLOPs & mIoU (\%)   \\
\toprule
PVTv2-B0\cite{wang2021pvtv2} & \textbf{7.6} & \textbf{25.0} & 37.2 \\
\bf \netname-XXS & 7.9 & 24.4 & \textbf{39.7} \\
\midrule
\bf \netname-XS & 10.6 & 27.7 & 41.4 \\
\midrule
ResNet18~\cite{resnet2016} & \textbf{15.5} & 32.2 & 32.9 \\
PVTv1-Tiny~\cite{pvt} & 17.0 & 33.2 & {35.7} \\
PVTv2-B1\cite{wang2021pvtv2} & 17.8 & 34.2 & 42.5 \\
\bf \netname-S & 16.9 & \bf 32.1 & \textbf{45.9} \\
\bottomrule
\end{tabular}
}
	\caption{\textbf{Semantic segmentation results on the validation set of ADE20K. 
	}
	{\em Segmentation model}: Semantic FPN \cite{semantic_fpn}.
	{\em GFLOPs}: Calculated at $512\times 512$ input size.
	}
	\label{tab:seg}
\end{table*}

\section{Conclusion}
\label{sec:conclusion}

In this work, we investigate the design of efficient ViTs
from the on-device deployment perspective.
By introducing a novel decomposition of self-attention,
we present a family of \netsname~that, for the first time, achieve comparable or even superior 
accuracy-efficiency tradeoff on generic visual recognition tasks, in comparison to 
a variety of state-of-the-art efficient CNNs and ViTs.
We conduct extensive on-device experiments 
using practically critical and previously underestimated metrics (e.g., energy-aware efficiency) 
and reveal new insights and observations in the comparison of light-weight CNN and ViT models.

\noindent {\bf Acknowledgements.} We thank Victor Escorcia, Yassine Ouali and Javier Fernandez for helpful discussions.

\bibliographystyle{splncs04}
\bibliography{mvit}
\clearpage
\appendix
\section{Appendix}
\label{sec:Appendix}
\subsection{Computing Complexity}
We calculate the computational cost of spatial context modeling involved in our proposed \shortmodulename~bottleneck. We omit point-wise operations for simplicity as the key difference is on the spatial modeling part.
Let us assume an input $X \in \mathcal{R}^{h\times w \times c}$ 
where $h$, $w$, $c$ denotes the height, the width, and the channel dimension, respectively.
The cost of the local aggregation is $\mathcal{O}(k^2hwc)$,
where $k^2$ is the local group size. 
By selecting one delegate out of $r^2$ tokens with $r$ the sub-sampling rate, the complexity of our Sparse Global Self-Attention is then 
$\mathcal{O}(\frac{h^2w^2}{r^4}c)$.
Finally, the local propagation step takes a cost of $\mathcal{O}(r^2hwc)$. Putting all these together we have a total cost of \shortmodulename~ is 
$\mathcal{O}( k^{2}hwc + \frac{ h^2w^2}{r^4}c  + r^{2}hwc)$.
When comparing with the cost of a standard multi-head self-attention 
$\mathcal{O}(h^2w^2c)$, 
we can see that our \shortmodulename~significantly reduces the computation overhead when $k \ll h, w$; and  $r > 1$. In our experiments, for simplicity we set $k=3$, and $r$ to (4,2,2,1) for the four stages.

\subsection{Implementation Details}

All variants of \netsname~can be built upon these components according to the schematic overview  (Fig. 2a of the main paper), and the model configuration parameters (Table 1a. of the main paper).
For more details with the ablation studies in the main paper, 
we have replaced or removed one of these blocks with the details given below. 

(1) In Table 4a, for the case of w/o LA, we aim to test the importance of separate local and global context modeling. Thus we remove both \texttt{LocalAgg} and \texttt{GlobalSparseAttn}, and instead use the Spatial-Reduced Self-Attention\footnote{\href{https://github.com/whai362/PVT/blob/v2/classification/pvt_v2.py\#L54-L126}{https://github.com/whai362/PVT/blob/v2/classification/pvt\_v2.py\#L54-L126}} introduced in PVT~\cite{wang2021pvtv2}, resulting in a single Self-Attention Block for both local and global context modeling. For the case of LA(LSA) we simply replace \texttt{LocalAgg} with Local-grouped Self-Attention\footnote{\href{https://github.com/Meituan-AutoML/Twins/blob/main/gvt.py\#L32-L71}{https://github.com/Meituan-AutoML/Twins/blob/main/gvt.py\#L32-L71}}  introduced in~\cite{chu2021twins}. 

(2) In Table 4b, 
we replace the default sampler ({\bf Center})  with {\bf Avg} and {\bf Max} functions which can be implemented with \texttt{AvgPool2d()} and \texttt{MaxPool2d()} in Pytorch~\cite{pytorch}, respectively. Note, for both cases the kernel size is set to \texttt{sample\_rate}. 

(3) In Table 4c, in the case of w/o LP, we replace our \texttt{GlobalSparseAttn} with Spatial-Reduce Self-Attention from PVT~\cite{wang2021pvtv2}, but different from w/o LA, we keep the \texttt{LocalAgg}. For the case of LP(Bilinear), the \texttt{LocalProp} is instantiated as a bilinear interpolation function (\texttt{Upsample(mode=`bilinear')} in Pytorch).

Note, the number of layers for each of these variants is down-scaled to have~0.5GFLOPs for fair comparison.

\subsection{Accuracy-Speed Pareto-Optimal Models}
In order to facilitate the Accuracy vs. Speed interpretation. We identify pareto optimal models when comparing trade-off between accuracy and latency \cite{pareto}. In our context, the accuracy-latency pareto-optimal models are defined as those upon which no other models can improve in either accuracy or latency without degrading other metrics. As shown in Fig.~\ref{fig:fig1}, our \netsname~are well comparable with best efficient CNNs \cite{mobilenetv2_cvpr18,mobilenetv3_iccv19,tan2019efficientnet}, whilst significantly dominating over all prior ViT counterparts. 
Specifically, \netsname{} are all pareto-optimal 
in both trade-offs.

\begin{figure}
    \centering
    \includegraphics[width=1\linewidth]{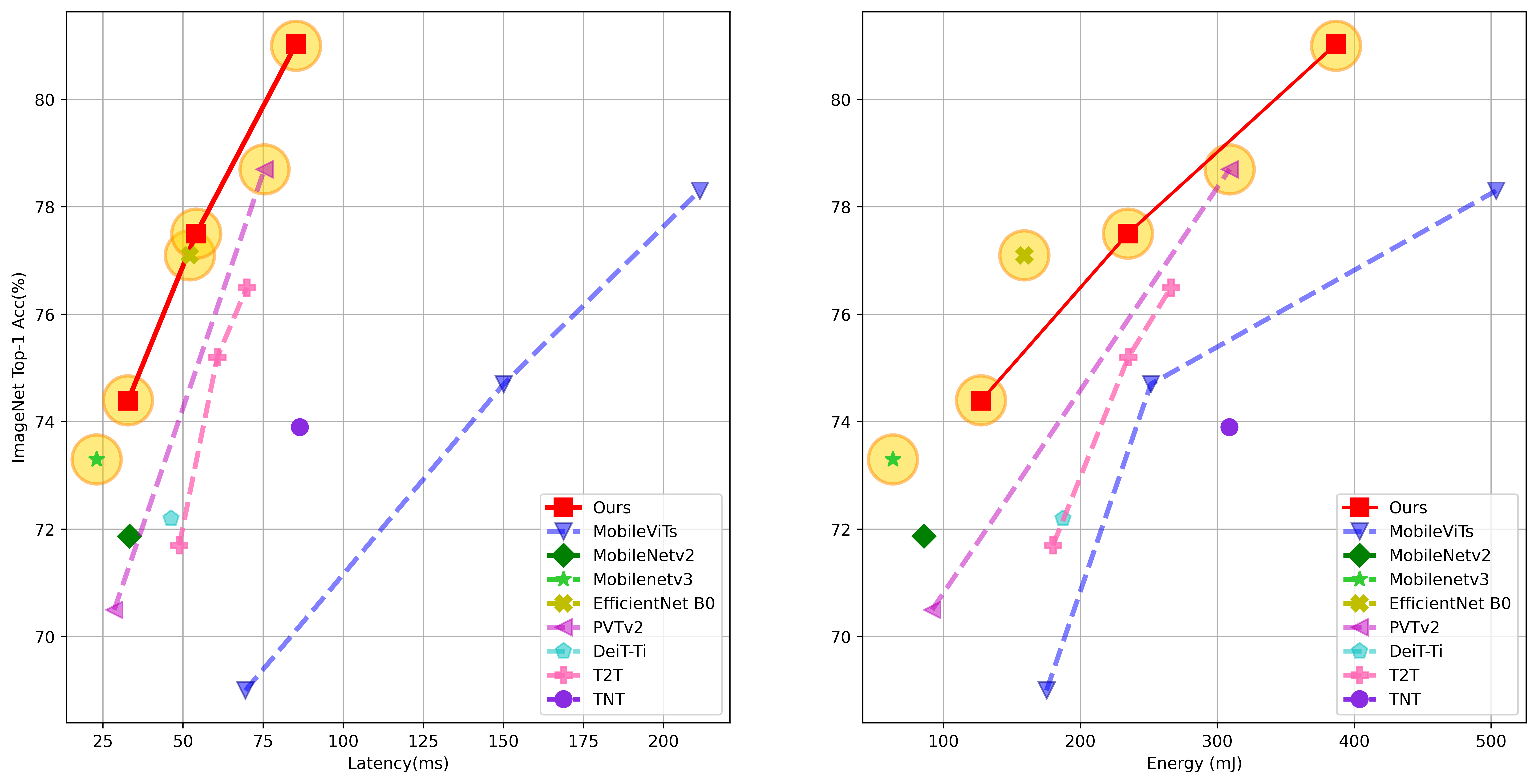}
    \caption{ {\bf Accuracy/Latency and Accuracy/Energy trade off on ImageNet-1K.} Note that, all three variants of our \netsname~are pareto-optimal, which are highlighted with {\color{amber}
    amber cercle}.
    {\em Testing device}: Samsung Galaxy S21 (latency), Snapdragon 888 Hardware Development Kit (energy).
    }
    \label{fig:fig1_appendix}
\end{figure}

\subsection{Efficiency in detection/segmentation} \label{sec:dense_efficiency}
This work proposes a genetic transformer-based network and demonstrate its efficacy when used as the backbone in detection/segmentation. We provide a evaluation by measuring only the inference time, energy and efficiency of the backbones for detection/segmentation. As shown in Tab.~\ref{tab:detection_eff} and \ref{tab:segmentation_eff}, EdgeViTs demonstrate higher efficiency compared to the baselines. 
\begin{table}[!t]
\centering
\begin{tabular}{@{}l|x{30}|x{35}|x{38}|x{40}@{}}
\toprule
Model & AP (\%) & CPU (s) & Energy (J) & Efficiency (\%/msW) \\
\toprule
{PVTv2-B0} & 37.2 & 1.34{\tiny$\pm$0.05}  & 3.50{\tiny$\pm$0.77} & 0.011  \\
{ResNet18} & 31.8 & 0.58{\tiny$\pm$0.02}  & 2.22{\tiny$\pm$0.35} & 0.014  \\
{PVTv1-Tiny} & 36.7 & 1.91{\tiny$\pm$0.18} & 4.40{\tiny$\pm$0.94} & 0.008  \\
{PVTv2-B1} & 41.2 & 2.81{\tiny$\pm$0.26} & 5.29{\tiny$\pm$1.49} & 0.008  \\
\midrule
{\bf \netname-XXS} & 38.7  & 0.59{\tiny$\pm$0.02} & 2.02{\tiny$\pm$0.58} & 0.019 \\
{\bf \netname-XS} & 40.6  & 0.90{\tiny$\pm$0.03}& 2.89{\tiny$\pm$0.66}& 0.014 \\
{\bf \netname-S} & 43.4 & 1.88{\tiny$\pm$0.05} & 4.36{\tiny$\pm$1.06} & 0.010 \\
\bottomrule
\end{tabular}
\caption{\textbf{Detection Efficiency.} Input size: $800\times800$
}
\label{tab:detection_eff}
\end{table}
\begin{table}[!t]
\centering
\begin{tabular}{@{}l|x{30}|x{35}|x{38}|x{40}@{}}
\toprule
Model & mIoU (\%) & CPU (s) & Energy (J) & Efficiency (\%/msW) \\
\toprule
{PVTv2-B0} & 37.2  & 0.35{\tiny$\pm$0.01}  & 1.10{\tiny$\pm$0.30} & 0.034  \\
{ResNet18} &  32.9 & 0.23{\tiny$\pm$0.01} & 1.03{\tiny$\pm$0.20} & 0.032  \\
{PVTv1-Tiny} & 35.7  & 0.49{\tiny$\pm$0.02} & 1.63{\tiny$\pm$0.32} & 0.022  \\
{PVTv2-B1} & 42.5 & 0.75{\tiny$\pm$0.03} & 2.13{\tiny$\pm$0.82} &  0.020 \\
\midrule
{\bf \netname-XXS} & 39.7  & 0.19{\tiny$\pm$0.01} & 0.71{\tiny$\pm$0.11}  & 0.056 \\
{\bf \netname-XS} & 41.4 & 0.31{\tiny$\pm$0.01} & 1.11{\tiny$\pm$0.28} & 0.037 \\
{\bf \netname-S} & 45.9 & 0.52{\tiny$\pm$0.02} & 1.73{\tiny$\pm$0.36} & 0.027 \\
\bottomrule
\end{tabular}
\caption{\textbf{Segmentation Efficiency.} Input size: $512\times512$ 
}
\label{tab:segmentation_eff}
\end{table}

\end{document}